\documentclass[10pt,twocolumn,letterpaper]{article}

\usepackage{iccv}
\usepackage{times}
\usepackage{epsfig}
\usepackage{graphicx}
\usepackage{amsmath}
\usepackage{amssymb}
\usepackage[ruled,vlined,linesnumbered]{algorithm2e}

\usepackage[breaklinks=true,bookmarks=false]{hyperref}

\iccvfinalcopy 

\def\iccvPaperID{3107} 

\ificcvfinal\fi

\begin{document}

\title{Unsupervised Multi-Task Feature Learning on Point Clouds}

\author{Kaveh Hassani\\
Autodesk AI Lab\\
Toronto, Canada\\
{\tt\small kaveh.hassani@autodesk.com}
\and
Mike Haley\\
Autodesk AI Lab\\
San Francisco, USA\\
{\tt\small mike.haley@autodesk.com}
}

\maketitle
\ificcvfinal\thispagestyle{empty}\fi

\begin{abstract}
   We introduce an unsupervised multi-task model to jointly learn point and shape 
   features on point clouds. We define three unsupervised tasks including clustering,
   reconstruction, and self-supervised classification to train a multi-scale graph-based 
   encoder. We evaluate our model on shape classification and segmentation 
   benchmarks. The results suggest that it outperforms prior state-of-the-art 
   unsupervised models: In the ModelNet40 classification task, it achieves an accuracy 
   of 89.1\% and in ShapeNet segmentation task, it achieves an mIoU of 68.2 and 
   accuracy of 88.6\%.
\end{abstract}

\section{Introduction}
Point clouds are sparse order-invariant sets of interacting points defined in a 
coordinate space and sampled from surface of objects to capture their 
spatial-semantic information. They are the output of 3D sensors such as LiDAR 
scanners and RGB-D cameras, and are used in applications such as human-computer interactions \cite{Ge_2018_CVPR}, self-driving cars \cite{Qi_2018_CVPR}, and robotics \cite{rusu_2008_RAS}. Their sparse nature makes them computationally efficient and 
less sensitive to noise compared to volumetric and multi-view representations.

Classic methods craft salient geometric features on point clouds to capture their local 
or global statistical properties. Intrinsic features such as wave kernel signature (WKS) \cite{Aubry_2011_ICCV}, heat kernel signature (HKS) \cite{Bronstein_2010_CVPR}, 
multi-scale Gaussian curvature \cite{Sun_2009_CGF}, and global point signature \cite{Rustamov_2007_EG}; and extrinsic features such as persistent point feature 
histograms \cite{Rusu_2008_IROS} and fast point feature histograms 
\cite{Rusu_2009_CRA} are examples of such features. These features cannot address semantic tasks required by modern applications and hence are replaced by the 
unparalleled representation capacity of deep models. 

Feeding point clouds to deep models, however, is not trivial. Standard deep models 
operate on regular-structured inputs such as grids (images and volumetric data) and sequences (speech and text) whereas point clouds are permutation-invariant and 
irregular in nature. One can rasterize the point clouds into voxels \cite{Wu_CVPR_2015, Maturana_IROS_2015, Qi_2016_CVPR} but it demands excessive time and memory, and suffers from information loss and quantization artifacts \cite{Wang_2018_ArXiv}. 

Some recent deep models can directly consume point clouds and learn to perform 
various tasks such as classification \cite{Wang_2018_ArXiv}, semantic segmentation \cite{Ye_2018_ECCV, Engelmann_2017_ICCV}, part segmentation 
\cite{Wang_2018_ArXiv}, image-point cloud translation \cite{Fan_2017_CVPR}, object detection and region proposal \cite{Zhou_2018_CVPR}, consolidation and surface reconstruction \cite{Yu_2018_ECCV, Mostegel_2017_CVPR, Nan_2017_ICCV}, 
registration \cite{Elbaz_2017_CVPR, Vongkulbhisal_2018_CVPR, Lawin_2018_CVPR}, 
generation \cite{Sun_2018_ArXiv, Li_2018_ArXiv}, and up-sampling \cite{Yu_2018_CVPR}. 
These models achieve promising results thanks to their feature learning capabilities. However, to successfully learn such features, they require large amounts of labeled 
data. 

A few works explore unsupervised feature learning on point sets using autoencoders \cite{Yang_2018_CVPR, Deng_2018_ECCV, Li_2018_CVPR, Zhao_2019_CVPR, Achlioptas_2018_ICLR, Elbaz_2017_CVPR} and generative models, e.g., generative 
adversarial networks (GAN) \cite{Sun_2018_ArXiv, Li_2018_ArXiv, 
Achlioptas_2018_ICLR}, variational autoencoders (VAE) \cite{Gadelha_2018_ECCV}, and Gaussian mixture models (GMM) \cite{Achlioptas_2018_ICLR}. Despite their good 
feature learning capabilities, they suffer from not having access to supervisory signals 
and targeting a single task. These shortcomings can be addressed by self-supervised learning and multi-task learning, respectively. Self-supervised learning defines a pretext 
task using only the information present in the data to provide a surrogate supervisory 
signal whereas multi-task learning uses the commonalities across tasks by jointly 
learning them \cite{Zhang_2017_arXiv}.

We introduce a multi-task model that exploits three regimes of unsupervised learning including self-supervision, autoencoding, and clustering as its target tasks to jointly 
learn point and shape features. Inspired by \cite{Caron_2018_ECCV, Dizaji_2017_ICCV}, 
we show that leveraging joint clustering and self-supervised classification along with enforcing reconstruction achieves promising results while avoiding trivial solutions. 
The key contributions of our work are as follows:

\begin{itemize}
\item We introduce a multi-scale graph-based encoder for point clouds and train it 
within an unsupervised multi-task learning setting.
\item We exhaustively evaluate our model under various learning settings on 
ModelNet40 shape classification and ShapeNetPart segmentation tasks.
\item We show that our model achieves state-of-the-art results w.r.t prior 
unsupervised models and narrows the gap between unsupervised and supervised 
models.
\end{itemize}

\section{Related Work}
\subsection{Deep Learning on Point Clouds}

PointNet \cite{Qi_2017_CVPR} is an MLP that learns point features independently and aggregates them into a shape feature. PointNet++ \cite{Qi_2017_NIPS} defines 
multi-scale regions and uses PointNet to learn their features and then hierarchically aggregates them. Models based on KD-trees \cite{Klokov_2017_ICCV, 
Zeng_2018_ECCV, Gadelha_2018_ECCV}  spatially partition the points using kd-trees 
and then recursively aggregate them. RNNs \cite{Huang_2018_CVPR, Ye_2018_ECCV, Engelmann_2017_ICCV, Liu_2018_Arxiv} are applied to point clouds by the assumption 
that \textit{``order matters''} \cite{Vinyals_2015_ICLR} and achieve promising results 
on semantic segmentation tasks but the quality of the learned features is not clear.

CNN models introduce non-Euclidean convolutions to operate on point sets. A few 
models such as RGCNN \cite{Te_2018_MM}, SyncSpecCNN \cite{Yi_2017_CVPR} and 
Local Spectral GCNN \cite{Wang_2018_ECCV} operate on \emph{spectral} domain. 
These models tend to be computationally expensive. \emph{Spatial} CNNs learn point features by aggregating the contributions of neighbor points. Pointwise convolution \cite{Hua_2018_CVPR}, Edge convolution \cite{Wang_2018_ArXiv} , Spider convolution \cite{Xu_2018_ECCV}, sparse convolution \cite{Su_2018_CVPR, Graham_2018_CVPR}, 
Monte Carlo convolution \cite{Hermosilla_2018_TOG}, parametric continuous 
convolution \cite{Wang_2018_CVPR}, feature-steered graph convolution \cite{
Verma_2018_CVPR}, point-set convolution \cite{Shen_2018_CVPR}, $\chi$-convolution \cite{Li_2018_NIPS}, and spherical convolution \cite{Lei_2018_Arxiv} are examples of 
these models. Spatial models provide strong localized filters but struggle to learn 
global structures \cite{Te_2018_MM}.

A few works train generative models on point sets. Multiresolution VAE \cite{ Gadelha_2018_ECCV} introduces a VAE with multiresolution  convolution and 
deconvolution layers. PointGrow \cite{Sun_2018_ArXiv} is an auto-regressive model 
that can generate point clouds from scratch or conditioned on given semantic contexts. 
It is shown that GMMs trained on PointNet features achieve better performance 
compared to GANs \cite{Achlioptas_2018_ICLR}. 

A few recent works explore representation learning using autoencoders. A simple autoencoder based on PointNet is shown to achieve good results on various tasks \cite{Achlioptas_2018_ICLR}. FoldingNet \cite{Yang_2018_CVPR} uses an encoder with 
graph pooling and MLP layers and introduces a decoder of folding operations that 
deform a 2D grid onto the underlying object surface. PPF-FoldNet \cite{
Deng_2018_ECCV} projects the points into point pair feature (PPF) space and then 
applies a PointNet encoder and a FoldingNet decoder to reconstruct that space. 
AtlasNet\cite{Groueix_2018_CVPR} extends the FoldingNet to multiple grid patches 
whereas SO-Net \cite{Li_2018_CVPR} aggregates the point features into SOM node 
features to encode the spatial distributions. PointCapsNet \cite{Zhao_2019_CVPR} 
introduces an autoencoder based on dynamic routing to extract latent capsules and 
a few MLPs that generate multiple point patches from the latent capsules with distinct 
grids. 

\subsection{Self-Supervised Learning}

Self-supervised learning defines a proxy task on unlabeled data and uses the 
pseudo-labels of that task to provide the model with supervisory signals. It is used in 
machine vision with proxy tasks such as predicting arrow of time \cite{Wei_2018_CVPR}, missing pixels \cite{Pathak_2016_CVPR}, position of patches \cite{Doersch_2015_ICCV}, 
image rotations \cite{Gidaris_2018_ICLR}, synthetic artifacts \cite{Jenni_2018_CVPR}, 
image clusters \cite{Caron_2018_ECCV}, camera transformation in consecutive frames \cite{Agrawal_2015_ICCV}, rearranging shuffled patches \cite{Noroozi_2016_ECCV}, 
video colourization \cite{Vondrick_2018_ECCV}, and tracking of image patches\cite{ Wang_2015_ICCV} and has demonstrated promising results in learning and transferring 
visual features.

The main challenge in self-supervised learning is to define tasks that relate most to the down-stream tasks that use the learned features \cite{Jenni_2018_CVPR}. Unsupervised learning, e.g., density estimation and clustering, on the other hand, is not domain 
specific \cite{Caron_2018_ECCV}. \emph{Deep clustering} \cite{Aljalbout_2018_ArXiv, Min_2018_IA, Yang_2017_ICML, Hershey_2016_ICASSP, Xie_2016_ICML, 
Dizaji_2017_ICCV, Shaham_2018_ICLR, Yang_2016_CVPR, Hsu_2018_MUL} models are recently proposed to learn cluster-friendly features by jointly optimizing a clustering 
loss with a network-specific loss. A few recent works combine these two approaches 
and define deep clustering as a surrogate task for self-supervised learning. It is shown 
that alternating between clustering the latent representation and predicting the cluster assignments achieves state-of-the-art results in visual feature learning\cite{Caron_2018_ECCV, Dizaji_2017_ICCV}.

\subsection{Multi-Task Learning}
Multi-task learning leverages the commonalities across relevant tasks to enhance the performance over those tasks \cite{Zhang_2017_arXiv, Evgeniou_2007_NIPS}. It learns 
a shared feature with adequate expressive power to capture the useful information 
across the tasks. Multi-task learning has been successfully used in machine vision applications such as image classification \cite{Luo_2015_TIP}, image segmentation 
\cite{Dai_2016_CVPR}, video captioning \cite{Pasunuru_2017_ACL}, and activity 
recognition \cite{Yan_2015_TIP}. A few works explore self-supervised multi-task 
learning to learn high level visual features \cite{Doersch_2017_CVPR, Ren_2018_CVPR}. 
Our approach is relevant to these models except we use self-supervised tasks in 
addition to other unsupervised tasks such as clustering and autoencoding.  

\section{Methodology} \label{methodology}

Assume a training set $\mathcal{S}=[s_1,s_2,...,s_N]$ of $N$ point sets where a point 
set $s_i= \lbrace p_{1}^{i},p_{2}^{i},...,p_{M}^{i} \rbrace$ is an order-invariant set of $M$ points and each point $p_{j}^{i} \in \mathbb{R}^{d_{in}}$. In the simplest case $p_{j}^{i} = 
(x_{j}^{i}, y_{j}^{i}, z_{j}^{i})$ only contains coordinates, but can extend to carry other 
features, e.g., normals. We define an \textbf{encoder} $E_\theta: \mathcal{S} 
\longmapsto \mathcal{Z}$ that maps input point sets from $\mathbb{R}^{M \times 
d_{in}}$ into the latent space $\mathcal{Z} \in \mathbb{R}^{d_z}$ such that $d_z \gg 
d_{in}$. For each point $p_{j}^{i}$, the encoder first learns a point (local) feature 
$z_{j}^{i} \in \mathbb{R}^{d_z}$ and then aggregates them into a shape (global) feature 
$Z^{i} \in \mathbb{R}^{d_z}$. It basically projects the input points to a feature subspace 
with higher dimension to encode richer local information than the original space. Any parametric non-linear function parametrized by $\theta$ can be used as the encoder. 
To learn $\theta$ in unsupervised multi-task fashion, we define three parametric 
functions on the latent variable $Z$ as follows: 

\textbf{Clustering function} $\Gamma_c: \mathcal{Z} \longmapsto \mathcal{Y}$ maps 
the latent variable into $K$ categories $\mathcal{Y}=[y_1,y_2,...,y_n]$ such that $y_i \in \lbrace 0, 1\rbrace^K$ and $y_n^T\mathbf{1}_k=1$. This function encourages the 
encoder to generate features that are clustering-friendly by pushing similar samples in 
the feature space closer and pushing dissimilar ones away. It also provides the model 
with pseudo-labels for self-supervised learning through its hard cluster assignments.

\textbf{Classifier function} $f_\psi: \mathcal{Z} \longmapsto \mathcal{\hat{Y}}$ predicts 
the cluster assignments of the latent variable such that the predictions correspond to 
the hard clusters assignments of $\Gamma_c$. In other words, $f_\psi$ maps the latent variable into $K$ predicted categories $\mathcal{\hat{Y}}=[\hat{y}_1,\hat{y}_2,...,
\hat{y}_n]$ such that $\hat{y}_i \in \lbrace 0, 1\rbrace^K$. This function uses the 
pseudo-labels generated by the clustering function, i.e., cluster assignments, as its 
proxy train data. The difference between the cluster assignments and the predicted 
cluster assignments provides the supervisory signals.

\textbf{Decoder function} $g_\phi: \mathcal{Z} \longmapsto \mathcal{\hat{S}}$ 
reconstructs the original point set from the latent variable, i.e., maps the latent variable 
$\mathcal{Z} \in \mathbb{R}^{d_{z}}$ to a point set $\mathcal{\hat{S}} \in \mathbb{R}^
{M \times d_{in}}$. Training a deep model with a clustering loss collapses the features 
into a single cluster \cite{Caron_2018_ECCV}. Some heuristics such as penalizing the 
minimal number of points per cluster and randomly reassigning empty clusters are 
introduced to prevent this. We introduce the decoder function to prevent the model 
from converging to trivial solutions.

\subsection{Training}
The model alternates between clustering the latent variable $\mathcal{Z}$ to generate pseudo-labels $\mathcal{Y}$ for self-supervised learning, and learning the model 
parameters by jointly predicting the pseudo-labels $\mathcal{\hat{Y}}$ and 
reconstructing the input point set $\mathcal{\hat{S}}$. Assuming \emph{K-means} 
clustering, the model learns a centroid matrix $C \in \mathbb{R}^{d_z \times K}$ and 
cluster assignments $y_n$ by optimizing the following objective clustering \cite{Caron_2018_ECCV}:
\begin{equation}
\min_{\lbrace C, \theta \rbrace}{\frac{1}{N} \sum_{n=1}^{N}{\min_{y_n \in \lbrace 0, 
1\rbrace^K} \left\|z_n-Cy_n\right\|_{2}^{2}}}
\end{equation}
where $z_n=E_\theta(s_n)$ and $y_n^T\mathbf{1}_k=1$. The centroid matrix is 
initialized randomly. It is noteworthy that: (i) when assigning cluster labels, the centroid matrix is fixed, and (ii) the centroid matrix is updated epoch-wise and not batch-wise to prevent the learning process from diverging.

For the classification function, we minimize the cross-entropy loss between the cluster assignments and the predicted cluster assignments as follows.
\begin{equation}
\min_{\lbrace \theta, \psi \rbrace}{\frac{-1}{N}\sum_{n=1}^{N}{y_n\log{\hat{y}_n}}}
\end{equation}
where $y_n=\Gamma_c(z_n)$ and $\hat{y}_n=f_\psi(z_n)$ are the cluster assignments 
and the predicted cluster assignments, respectively.

We use Chamfer distance to measure the difference between the original point cloud 
and its reconstruction. Chamfer distance is differentiable with respect to points and is 
computationally efficient. It is computed by finding the nearest neighbor of each point 
of the original space in the reconstructed space and vice versa, and summing up their 
Euclidean distances. Hence, we optimize the decoding loss as follows.
\begin{equation}
\min_{\lbrace \theta, \phi \rbrace}
\frac{1}{2NM}
\sum_{n=1}^{N} \sum_{m=1}^{M}
\min_{\hat{p} \in \hat{s}_n} \left\|p_{m}^{n}-\hat{p}\right\|_{2}^{2} + 
\min_{p \in s_n} \left\|\hat{p}_{m}^{n}-p\right\|_{2}^{2}
\end{equation}
where $\hat{s}_n=g_\phi(z_n)$ and, $s_n$ and $\hat{s}_n$ are the original and 
reconstructed point sets, respectively. $N$ and $M$ denote the number of point 
sets in the train set and the number of points in each point set, respectively.

Let's denote the clustering, classification, and decoding objectives by $\mathcal{L}_{\Gamma}$, $\mathcal{L}_{f}$, and $\mathcal{L}_{g}$, respectively. we define the 
multi-task objective as a linear combination of these objectives: $\mathcal{L} = 
\alpha\mathcal{L}_{\Gamma} + \beta\mathcal{L}_{f} + \gamma\mathcal{L}_{g}$ and 
train the model based on that. The training process is shown in Algorithm \ref{algo}. 

We first randomly initialize the model parameters and assume an arbitrary upper 
bound for the number of clusters. We show through experiments that the model 
converges to a fixed number of clusters by emptying some of the clusters. This is 
especially favorable when the true number of categories is unknown. We then 
randomly select $K$ point sets from the training data and feed them to the 
randomly initialized encoder and set the extracted features as the initial centroids. 
Afterwards we optimize the model parameters w.r.t the multi-task objective using 
mini-batch stochastic gradient descent. Updating the centroids with the same 
frequency as the network parameters can destabilize the training. Therefore, we 
aggregate the learned features and the cluster assignments within each epoch 
and update the centroids after an epoch is completed.

\begin{algorithm}
\SetAlgoLined \DontPrintSemicolon
$\theta, \phi, \psi$ $\longleftarrow Random()$ 
\hfill\hfill\hfil\hfill\hfill Initial parameters\;
$K \longleftarrow K_{UB}$  
\hfill\hfill\hfil\hfill\hfill\hfill\hfill Upper bound  \#clusters\;
$C\longleftarrow E_\theta\left(Choice\left(\mathcal{S}, K\right)\right)$  
\hfill\hfill\hfill\hfil Initial centroids \; 
\For{epoch in epochs}{
	\While{epoch not completed}{
	\textbf{Forward pass} \;
	$\mathcal{S}_x\longleftarrow Sample(\mathcal{S})$ 
     \hfill\hfill\hfil\hfill\hfill  Mini-batch \;
	$\mathcal{Z}_x \longleftarrow E_\theta \left( \mathcal{S}_x \right)$ 
	\hfill\hfill\hfil\hfill\hfill  Encoding\;
	$\mathcal{Y}_x \longleftarrow \Gamma_c \left(\mathcal{Z}_x \right)$ 
	\hfill\hfill\hfil\hfill\hfill  Cluster assignment\;
	$\mathcal{\hat{Y}}_x \longleftarrow f_\psi \left(\mathcal{Z}_x \right)$ 
	\hfill\hfill\hfil\hfill\hfill  Cluster prediction \;
	$\mathcal{\hat{S}}_x \longleftarrow g_\phi \left(\mathcal{Z}_x \right)$ 
	\hfill\hfill\hfil\hfill\hfill  Decoding \;
	$\left(\mathcal{Z}, \mathcal{Y} \right) \longleftarrow Aggregate\left(\mathcal{Z}_x, \mathcal{Y}\right)$ \;
	\textbf{Backwards pass} \;
	
	$\bigtriangledown_{\theta, \phi, \psi} (\alpha\mathcal{L}_{\Gamma}(\mathcal{Z}_x, C; \theta) + $ 
	\hfill\hfill\hfil\hfill\hfill  Compute gradients \;
	$\beta\mathcal{L}_{f}(\mathcal{Y}_x, \mathcal{\hat{Y}}_x; \theta, \psi) +$ \;
	$\gamma\mathcal{L}_{g}(\mathcal{S}_x, \mathcal{\hat{S}}_x; \theta, \phi) )$ \;
	 $Update(\theta, \phi, \psi)$ 
	 \hfill\hfill\hfill\hfill\hfill\hfill\hfill\hfill Update with gradients \;
	}
$C\longleftarrow Update(\mathcal{Z}, \mathcal{Y})$ 
\hfill\hfill\hfill\hfill\hfill Update centroids \;
}
 \caption{Unsupervised Multi-task training algorithm.} \label{algo}
\end{algorithm}
	
\subsection{Architecture}
\begin{figure*}
\begin{center}
\includegraphics[width=1.0\linewidth]{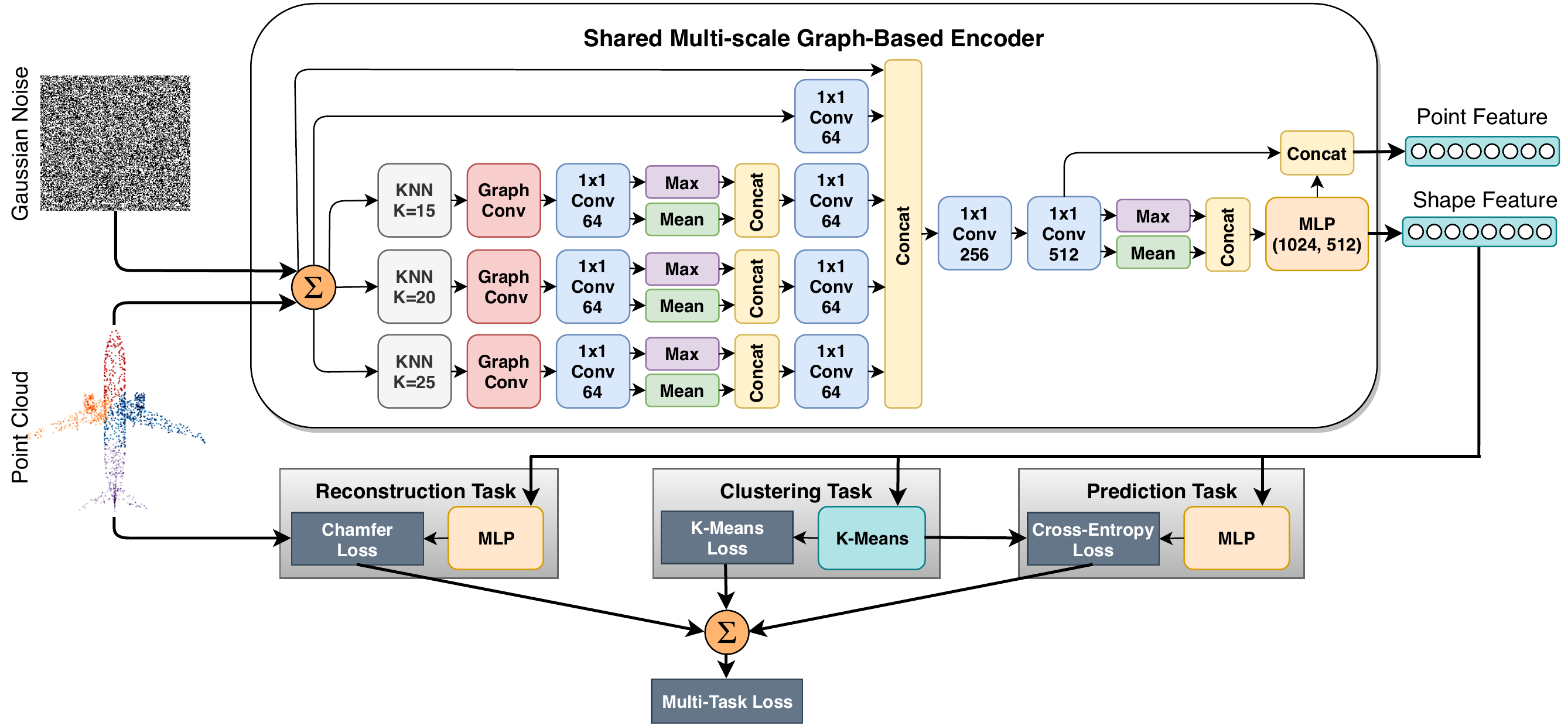}
\end{center}
   \caption{Proposed Architecture for unsupervised multi-task feature learning 
   on point clouds. It consists of a multi-scale graph-based encoder that generates
   point and shape features for an input point cloud and three task decoders that 
   jointly provide the architecture with a multi-task loss.}
\label{fig:arch}
\end{figure*}

Inspired by \emph{Inception} \cite{Szegedy_2015_CVPR} and Dynamic Graph CNN 
(DGCNN) \cite{Wang_2018_ArXiv} architectures, we introduce a graph-based 
architecture shown in Figure \ref{fig:arch} which consists of an encoder and three 
task-specific decoders. The encoder uses a series of graph convolution, 
convolution, and pooling layers in a multi-scale fashion to learn point and shape 
features from an input point cloud jittered by Gaussian noise. For each point, it 
extracts three intermediate features by applying graph convolutions on three 
neighborhood radii and concatenates them with the input point feature and its 
convolved feature. The first three features encode the interactions between each 
point and its neighbors where as the last two features encode the information about 
each point. The concatenation of the intermediate features is then passed through a 
few convolution and pooling layers to learn another level of intermediate features. 
These point-wise features are then pooled and fed to an MLP to learn the final shape 
feature. They are also concatenated with the shape feature to represent the final 
point features. Similar to \cite{Wang_2018_ArXiv}, we define the graph convolution 
as follows:

\begin{equation}
z_{i} = \sum_{p_k \in \mathcal{N}(p_i)} 
h_{\theta}\left(\left[ p_i \parallel p_k - p_i \right] \right)
\end{equation}

where $z_{i}$ is the learned feature for point $p_{i}$ based on its neighbor 
contributions, $p_k \in \mathcal{N}(p_i)$ are the $k$ nearest points to the $p_{i}$ 
in Euclidean space, $h_{\theta}$ is a nonlinear function parameterized by $\theta$ 
and $\parallel$ is the concatenation operator. We use a shared MLP for $h_{\theta}$. 
The reason to use both $p_i$ and $p_k - p_i$ is to encode both global information 
($p_i$) and local interactions ($p_k - p_i$) of each point.

To perform the target tasks, i.e., clustering, classification, and autoencoding, we 
use the following. For clustering, we use a standard implementation of K-means to
cluster the shape features. For self-supervised classification, we feed the shape 
features to an MLP to predict the category of the shape (i.e., cluster assignment 
by the clustering module). And for the autoencoding task, we use an MLP to 
reconstruct the original point cloud from the shape feature. This MLP is denoising 
and reconstructs the original point cloud before the addition of the Gaussian noise. 
All these models along with the encoder are trained jointly and end-to-end. Note 
that all these tasks are defined on the shape features. Because a shape feature is 
an aggregation of its corresponding point features, learning a good shape feature 
pushes the model to learn good point features too.

\section{Experiments} \label{experiments}
\subsection{Implementation Details} \label{details}

We optimize the network using Adam \cite{Kingma_2014_ICLR} with an initial learning 
rate of 0.003 and batch size of 40. The learning rate is scheduled to decrease by 0.8 
every 50 epochs. We apply batch-normalization \cite{Ioffe_2015_ICML} and ReLU 
activation to each layer and use dropout \cite{Srivastava_2014_JMLR} with $p=0.5$. 
To normalize the task weights to the same scale, we set the weights of clustering (
$\alpha$), classification ($\beta$), and reconstruction($\gamma$) to 0.005, 1.0, 500, respectively. For graph convolutions, we use neighborhood radii of 15, 20, and 25 (as suggested in \cite{Wang_2018_ArXiv}) and for normal convolutions we use 1$\times$1 kernels. We set the upper bound number of clusters ($K_{UB}$) to 500. We also set 
the size of the MLPs in prediction and reconstruction tasks to [2048, 1024, 500] and 
[2048, 1024, 6144], respectively. Note that the size of the last layers correspond to 
the upper bound number of clusters (500) and the reconstruction size (6144: 
2048$\times$3). Following \cite{Achlioptas_2018_ICLR} we set the shape and point 
feature sizes to 512 and 1024, respectively.

For preprocessing and augmentation we follow \cite{Qi_2017_CVPR,Wang_2018_ArXiv} 
and uniformly sample 2048 points and normalize them to a unit sphere. We also apply 
point-wise Gaussian noise of $N\sim(0, 0.01)$ and shape-wise random rotations 
between [-180, 180] degrees along $z$-axis and random rotations between [-20, +20] degrees along $x$ and $y$ axes.

The model is implemented with Tensorflow \cite{Abadi_2016_Tensorflow} on a Nvidia 
DGX-1 server with 8 Volta V100 GPUs. We used synchronous parallel training by 
distributing the training mini-batches over all GPUs and averaging the gradients to 
update the model parameters. With this setting, our model takes 830s on average to 
train one epoch on the ShapeNet (i.e, $\sim$55k samples of size 2048$\times$3). We 
train the model for 500 epochs. At test time, it takes 8ms on an input point cloud with 
size 2048$\times$3. 

\subsection{Pre-training for Transfer Learning}

Following the experimental protocol introduced in \cite{Achlioptas_2018_ICLR}, we 
pre-train the model across all categories of the ShapeNet dataset \cite{
Chang_2015_Arxiv} (i.e., 57,000 models across 55 categories) , and then transfer the 
trained model to two down-stream tasks including shape classification and part segmentation. After pre-training the model, we freeze its weights and do not fine-tune 
it for the down-stream tasks.

Following \cite{Caron_2018_ECCV}, we use Normalized Mutual Information (NMI) to 
measure the correlation between cluster assignments and the categories without 
leaking the category information to the model. This measure gives insight on the 
capability of the model in predicting category level information without observing the 
ground-truth labels. The model reaches an NMI of 0.68 and 0.62 on the train and 
validation sets, respectively which suggests that the learned features are 
progressively encoding category-wise information. 

We also observe that the model converges to 88 clusters (from the initial 500 
clusters) which is 33 more clusters compared to the number of ShapeNet categories. 
This is consistent with the observation that ``\textit{some amount of over-segmentation 
is beneficial}'' \cite{Caron_2018_ECCV}. The model empties more than 80\% of the 
clusters  but does not converge to the trivial solution of one cluster. We also trained 
our model on the 10 largest ShapeNet categories to investigate the clustering behavior 
where the model converged to 17 clusters. This confirms that model converges to a 
fixed number of clusters which is less than the initial upper bound assumption and is 
more than the actual number of categories in the data.

\begin{figure*}[t]
\begin{center}
   \includegraphics[width=0.9\linewidth]{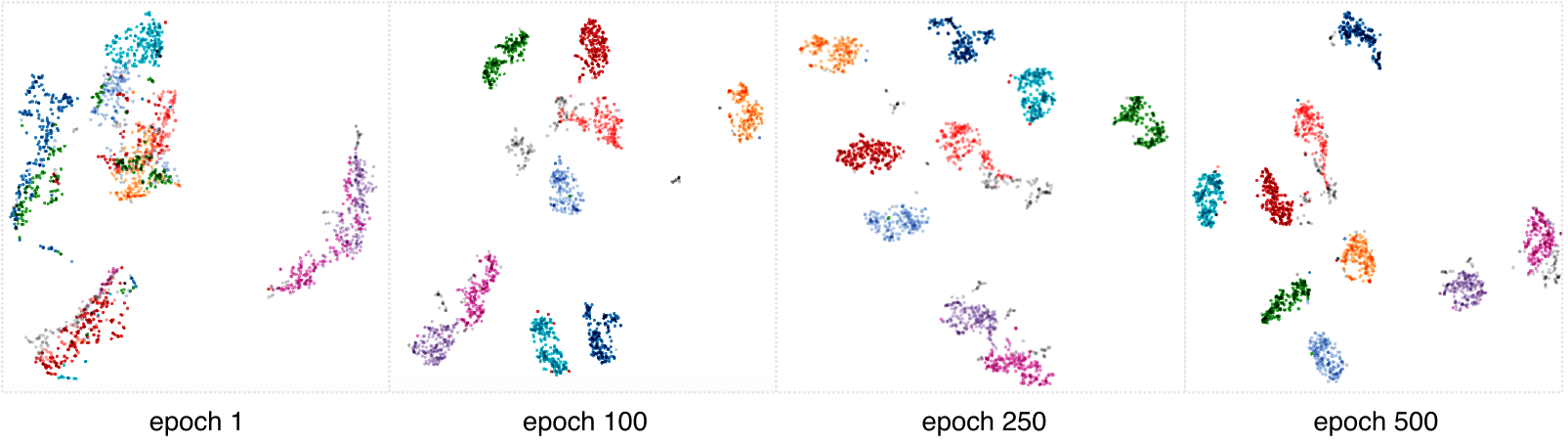}
\end{center}
   \caption{Evolution of the learned features along the training epochs 
   (visualized using t-SNE) showing progressive \emph{clustering-friendly} behavior. }
\label{fig:dynamics}
\end{figure*}

To investigate the dynamics of the learned features, we selected the 10 largest 
ShapeNet categories and randomly sampled 200 shapes from each category. The 
evolution of the features of the sampled shapes visualized using t-SNE (Figure \ref{fig:dynamics}) suggests that the learned features progressively demonstrate \emph{clustering-friendly} behavior along the training epochs. 

\subsection{Shape Classification}\label{sec:shpaecls}
To evaluate the performance of the model on shape feature learning, we follow the experimental protocol in \cite{Achlioptas_2018_ICLR} and report the classification 
accuracy on transfer learning from the ShapeNet dataset \cite{Chang_2015_Arxiv} 
to the ModelNet40 dataset \cite{Wu_2015_CVPR} (i.e., 13,834 models across 40 
categories divided to 9,843 and 3,991 train and test samples, respectively). Similar 
to \cite{Achlioptas_2018_ICLR}, we extract the shape features of the ModelNet40 
samples from the pre-trained model without any fine-tuning, train a linear SVM on 
them, and report the classification accuracy. This approach is a common practice in evaluating unsupervised visual feature learning \cite{Caron_2018_ECCV} and provides 
insight about the effectiveness of the learned features in classification tasks. 

Results shown in Table \ref{table2} suggest that our model achieves state-of-the-art accuracy on the ModelNet40 shape classification task compared to other unsupervised 
feature learning models. It is noteworthy that the reported result is without any hyper-parameter tuning. With random hyper-parameter search, we observed an 0.4 absolute increase in the accuracy (i.e., \textbf{89.5\%}). The results also suggest that the 
unsupervised model is competitive with the supervised models. Error analysis reveals 
that the misclassifications occur between geometrically similar shapes. For example, 
the three most frequent misclassifications are between (table, desk), (nightstand, 
dresser), and (flowerpot, plant) categories. A similar observation is reported in \cite{Achlioptas_2018_ICLR}  and it is suggested that stronger supervision signals 
may be required to learn subtle details that discriminate these categories. 

To further investigate the quality of the learned shape features, we evaluated them in a 
zero-shot setting. For this purpose, we cluster the learned features using agglomerative hierarchical clustering (AHC) \cite{Mullner_2011_Arxiv} and then align the assigned 
cluster labels with the ground truth labels (ModelNet40 categories) based on majority 
voting within each cluster. The results suggest that the model achieves \textbf{68.88\%} accuracy on the shape classification task with zero supervision. This result is consistent 
with the observed NMI between cluster assignments and ground truth labels in the 
ShapeNet dataset. 

\begin{table}
\addtolength{\tabcolsep}{-3pt}
\small
\begin{center}
\begin{tabular}{lc|lc}
\hline
\multicolumn{2}{c}{\textbf{Unsupervised transfer learning}} & \multicolumn{2}{c}{\textbf{Supervised learning}} \\
Model & Accuracy & Model & Accuracy\\
\hline\hline
SPH \cite{Kazhdan_2003_EG} & 68.2 & PointNet \cite{Qi_2017_CVPR} & 89.2\\
LFD \cite{Chen_2003_CGF} & 75.5 & PointNet++ \cite{Qi_2017_NIPS} & 90.7\\
T-L Network \cite{Girdhar_2016_ECCV}  & 74.4 & PointCNN \cite{Hua_2018_CVPR} & 86.1 \\
VConv-DAE \cite{Sharma_2016_ECCV} & 75.5 & DGCNN \cite{Wang_2018_ArXiv} & \textbf{92.2}\\
3D-GAN \cite{Wu_2016_NIPS} & 83.3 & KCNet \cite{Shen_2018_CVPR} & 91.0\\
Latent-GAN \cite{Achlioptas_2018_ICLR} & 85.7 & KDNet \cite{Klokov_2017_ICCV} & 91.8\\
MRTNet-VAE \cite{Gadelha_2018_ECCV} & 86.4 & MRTNet \cite{Gadelha_2018_ECCV} & 91.7\\
FoldingNet \cite{Yang_2018_CVPR} & 88.4 & SpecGCN \cite{Wang_2018_ECCV} & 91.5\\
PointCapsNet \cite{Zhao_2019_CVPR} & 88.9 & ~ & ~ \\
Ours & \textbf{89.1} \\
\hline
\end{tabular}
\end{center}
\caption{\textbf{Left}: Accuracy of classification by transfer learning from the ShapeNet 
on the ModelNet40 data. \textbf{Right}:  Classification accuracy of supervised learning 
on the ModelNet40 data. Our model narrows the gap with supervised models.} 
\label{table2}
\end{table}

\subsection{Part Segmentation}
Part segmentation is a fine-grained point-wise classification task where the goal is to 
predict the part category label of each point in a given shape. We evaluate the learned 
point features on the ShapeNetPart dataset \cite{Yi_2016_ATG}, which contains 16,881 objects from 16 categories (12149 train, 2874 test, and 1858 validation). Each object 
consists of 2 to 6 parts with total of 50 distinct parts among all categories. Following \cite{Qi_2017_CVPR}, we use mean Intersection-over-Union (mIoU) as the evaluation 
metric computed by averaging the IoUs of different parts occurring in a shape. We 
also report part classification accuracy. 

Following \cite{Zhao_2019_CVPR}, we randomly sample 1\% and 5\% of the 
ShapeNetPart train set to evaluate the point features in a semi-supervised setting.
We use the same pre-trained model to extract the point features of the sampled training
data, along with validation and test samples without any fine-tuning. We then train a
4-layer MLP [2048, 4096, 1024, 50] on the sampled training sets and evaluate it on all
test data. Results shown in Table \ref{table3} suggest that our model achieves 
state-of-the-art accuracy and mIoU on ShapeNetPart segmentation task compared to 
other unsupervised feature learning models. Also comparisons between our model
(trained on 5\% of the training data) and the fully supervised models are shown in Table 
\ref{table4}. The results suggest that our model achieves an mIoU which is only 8\% less 
than the best supervised model and hence narrows the gap with supervised models.

\begin{table}
\small
\begin{center}
\begin{tabular}{lcccc}
\hline

\textbf{Model} & \multicolumn{2}{c}{1\% of train data} & 
\multicolumn{2}{c}{5\% of train data} \\
\cline{2-5}
~ & \textbf{Accuracy} & \textbf{IoU} & \textbf{Accuracy} & \textbf{IoU}\\
\hline\hline
SO-Net\cite{Li_2018_CVPR} & 78.0 & 64.0 & 84.0 & 69.0\\
PointCapsNet\cite{Zhao_2019_CVPR} & 85.0 & 67.0 & 86.0 & 70.0\\
Ours & \textbf{88.6} & \textbf{68.2} & \textbf{93.7} & \textbf{77.7}\\
\hline
\end{tabular}
\end{center}
\caption{Results on semi-supervised ShapeNetPart segmentation task.}\label{table3}
\end{table}

\begin{table*}
\addtolength{\tabcolsep}{-5pt}
\small
\begin{center}
\begin{tabular}{lc|cc|cccccccccccccccc}
\hline

Model & \%train & Cat. & Ins. & Aero & Bag & Cap & Car & Chair & Ear & Guitar & Knife & Lamp & Laptop & Motor & Mug & Pistol & Rocket & Skate &	Table \\
~ & data & mIoU & mIoU & ~ & ~ & ~ & ~ & ~ & phone & ~ & ~ & ~ & ~ & ~ & ~ & ~ & ~ & board & ~ \\
\hline\hline
PointNet \cite{Qi_2017_CVPR} & ~ & 80.4 & 83.7 & 83.4 & 78.7 & 82.5 & 74.9 & 
89.6 & 73.0 & 91.5 & 85.9 & 80.8 & 95.3 & 65.2 & 93.0 & 81.2 & 57.9 & 72.8 & 80.6\\
PointNet++ \cite{Qi_2017_NIPS} & ~ & 81.9 & 85.1 & 82.4 & 79.0 & 87.7 & 77.3 & 90.8 
& 71.8 & 91.0 & 85.9 & 83.7 & 95.3 & 71.6 & 94.1 & 81.3 & 58.7 & 76.4 & 82.6 \\
DGCNN \cite{Wang_2018_ArXiv} & ~ & 82.3 & 85.1 & \textbf{84.2} & 83.7 & 84.4 & 77.1 
& \textbf{90.9} & \textbf{78.5} & 91.5 & 87.3 & 82.9 & 96.0 & 67.8 & 93.3 & 82.6 & 59.7 
& 75.5 & 82.0 \\
KCNet \cite{Shen_2018_CVPR} & ~ & 82.2 & 84.7 & 82.8 & 81.5 & 86.4 & 77.6 & 90.3 
& 76.8 & 91.0 & 87.2 & 84.5 & 95.5 & 69.2 & 94.4 & 81.6 & 60.1 & 75.2 & 81.3 \\
RSNet \cite{Huang_2018_CVPR} & ~ & 81.4 & 84.9 & 82.7& \textbf{86.4} & 84.1& 78.2 
& 90.4 & 69.3 & 91.4 & 87.0 & 83.5 & 95.4 & 66.0 & 92.6 & 81.8 & 56.1 & 75.8 & 82.2 \\
SynSpecCNN \cite{Yi_2017_CVPR} & \textbf{100\%}  & 82.0 & 84.7 & 81.6 & 81.7 & 81.9 
& 75.2 & 90.2 & 74.9 & 93.0 & 86.1 & \textbf{84.7} & 95.6 & 66.7 & 92.7&  81.6 & 60.6 & 82.9 & 82.1 \\
RGCNN \cite{Te_2018_MM} & ~ & 79.5 & 84.3 & 80.2 & 82.8 & \textbf{92.6} & 75.3 & 
89.2 & 73.7 & 91.3 & \textbf{88.4} & 83.3 & 96.0 & 63.9 & 95.7 & 60.9 & 44.6 & 72.9 
& 80.4 \\
SpiderCNN \cite{Xu_2018_ECCV} & ~ & 82.4 & 85.3 & 83.5 & 81.0 & 87.2 & 77.5 & 
90.7 & 76.8 & 91.1 & 87.3 & 83.3 & 95.8 & 70.2 & 93.5 & 82.7 & 59.7 & 75.8 & 
\textbf{82.8}\\
SPLATNet \cite{Su_2018_CVPR} & ~ & 83.7 & \textbf{85.4} & 83.2 & 84.3 & 89.1 & 
80.3 & 90.7 & 75.5 & 92.1 & 87.1 & 83.9 & 96.3 & 75.6 & \textbf{95.8} & 83.8 & 64.0 
& 75.5 & 81.8 \\
FCPN \cite{Rethage_2018_ECCV} & ~ & \textbf{84.0} & 84.0 & 84.0 & 82.8 & 86.4 & \textbf{88.3} & 83.3 & 73.6 & \textbf{93.4} & 87.4 & 77.4 & \textbf{97.7} & \textbf{81.4} 
& \textbf{95.8} & \textbf{87.7} & \textbf{68.4} & \textbf{83.6} & 73.4 \\
\hline
Ours & \textbf{5\%} & 72.1 & 77.7 & 78.4 & 67.7 & 78.2 & 66.2 & 85.5 & 52.6 & 87.7 & 
81.6 & 76.3 & 93.7 & 56.1 & 80.1 & 70.9 & 44.7 & 60.7 & 73.0 \\
\hline
\end{tabular}
\end{center}
\caption{Comparison between our semi-supervised model and supervised models on ShapeNetPart segmentation task. Average mIoU over instances (Ins.) and categories 
(Cat.) are reported.}\label{table4}
\end{table*}

\begin{figure}[t]
\begin{center}
   \includegraphics[width=1.0\linewidth]{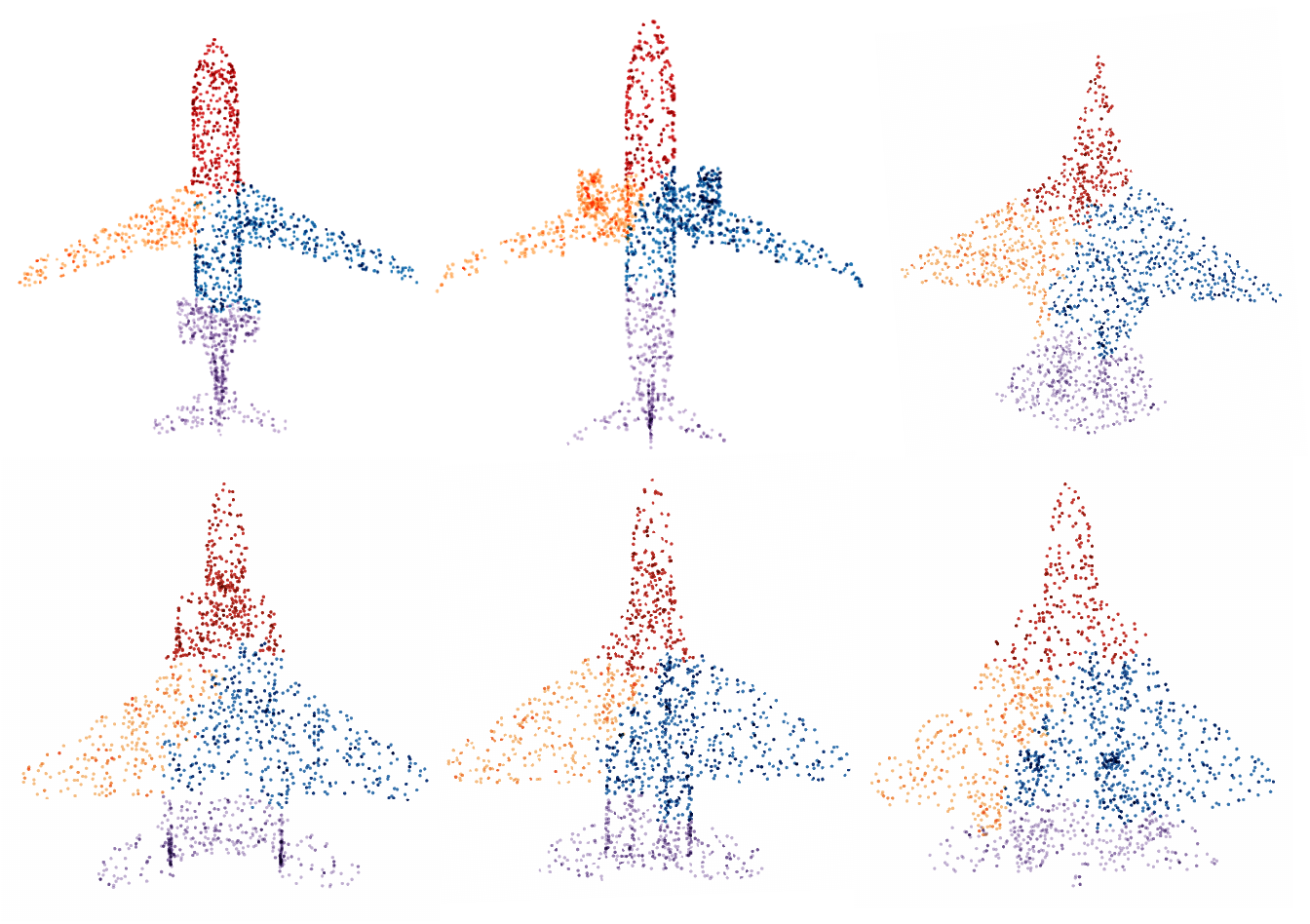}
\end{center}
   \caption{Co-clustering of the learned point features within the Airplane category 
   using hierarchical clustering which demonstrates the consistency of the learned 
   point features within the category.}
\label{fig:cocluster}
\end{figure}

\begin{figure*}[t]
\begin{center}
   \includegraphics[width=1.0\linewidth]{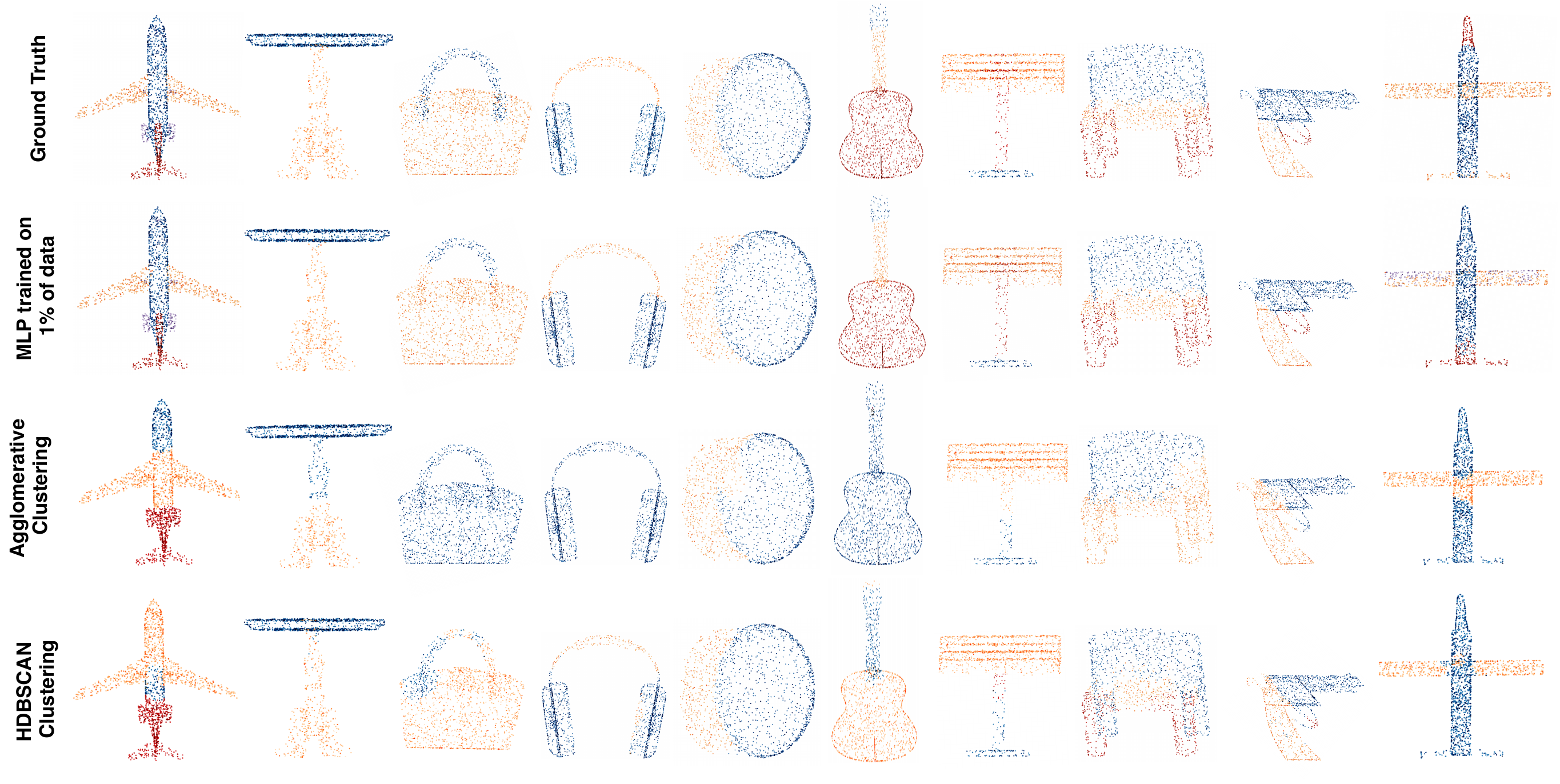}
\end{center}
   \caption{A few sample shapes along with their ground truth part labels, predicted 
   part labels by the trained MLP on 1\% of the training data, and predicted part labels 
   by AHC and HDBSCAN methods.}
\label{fig:seg}
\end{figure*}

We also performed intrinsic evaluations to investigate the consistency of the learned 
point features within each category. We sampled a few shapes from each category, 
stacked their point features, and reduced the feature dimension from 1024 to 512 
using PCA. We then co-clustered the features using the AHC method. The result of 
co-clustering on the \emph{airplane} category is shown in Figure \ref{fig:cocluster}. 
We observed a similar consistent behavior over all categories. We also used AHC and hierarchical density-based spatial clustering (HDBSCAN) \cite{Campello_2013_KDD}  methods to cluster the point features of each shape. We aligned the assigned cluster 
labels with the ground truth labels based on majority voting within each cluster. A few 
sample shapes along with their ground truth part labels, predicted part labels by the 
trained MLP, AHC, and HDBSCAN clustering are illustrated in Figure \ref{fig:seg}. As 
shown, HDBSCAN clustering results in a decent segmentation of the learned features 
in a fully unsupervised setting.
 
\subsection{Ablation Study}
We first investigate the effectiveness of the graph-based encoder on the shape 
classification task. In the first experiment, we replace the encoder with a PointNet \cite{Qi_2017_CVPR} encoder and keep the multi-task decoders. We train and test the network with the same transfer learning protocol which results in a classification 
accuracy of 86.2\%. Compared to the graph-based encoder with accuracy of 89.1\%, 
this suggests that our encoder learns better features and hence contributes to the 
state-of-the-art results that we achieve. To investigate the effectiveness of the 
multi-task learning, we compare our result against the results reported on a PointNet autoencoder (i.e., single reconstruction decoder) \cite{Achlioptas_2018_ICLR} which 
achieves classification accuracy of 85.7\%. This suggests that using multi-task learning improves the quality of the learned features. The summary of the results is shown in 
Table \ref{table6}.

\begin{table}
\small
\begin{center}
\begin{tabular}{cccc}
\hline
\textbf{Encoder} & \textbf{Decoder} & \textbf{Accuracy}\\
\hline\hline
PointNet & Reconstruction &  85.7 \\
PointNet & Multi-Task  & 86.2 \\
Ours & Reconstruction  & 86.7 \\
Ours & Multi-Task  & \textbf{89.1} \\
\hline
\end{tabular}
\end{center}
\caption{Effect of encoder and multi-task learning on accuracy on the ModelNet40.}\label{table6}
\end{table}

We also investigate the effect of different tasks on the quality of the learned features 
by masking the task losses and training and testing the model on each configuration. 
The results shown in Table \ref{table5} suggest that the reconstruction task has the 
highest impact on the performance. This is because contrary to \cite{
Caron_2018_ECCV}, we are not applying any heuristics to avoid trivial solutions and 
hence when the reconstruction task is masked both clustering and classification tasks 
tend to collapse the features to one cluster which results in degraded feature learning.

Moreover, the results suggest that masking the cross-entropy loss degrades the 
accuracy to 87.6\% (absolute decrease of 1.5\%) whereas masking the k-means loss
has a less adverse effect (degraded loss of 88.3\%, i.e., absolute decrease of 0.8\%). 
This implies that the cross-entropy loss (classifier) plays a more important role than 
the clustering loss. Furthermore, the results indicate that having both K-means and
cross-entropy losses along with the reconstruction task yields the best result (i.e., 
accuracy of 89.1\%). This may seems counter-intuitive as one may assume that using 
the clustering pseudo-labels to learn a classification function would push the classifier
to replicate the K-means behavior and hence the k-means loss will be redundant.
However, we think this is not the case because the classifier introduces non-linearity 
to the feature space by learning non-linear boundaries to approximate the predictions 
of the linear K-means model which in turn affects the clustering outcomes in the 
following epoch. K-means loss on the other hand, pushes the features in the same 
cluster to a closer space while pushing the features of other clusters away. 

\begin{table}
\addtolength{\tabcolsep}{-5pt}
\small
\begin{center}
\begin{tabular}{cccc}
\hline
\textbf{Classification} & \textbf{Reconstruction} & \textbf{Clustering} & \textbf{Overall}\\
\textbf{Task} & \textbf{Task} & \textbf{Task} & \textbf{Accuracy}\\
\hline\hline
$\surd$ & $\times$ & $\times$ &  22.8 \\
$\times$ & $\surd$  & $\times$ & 86.7 \\
$\times$ & $\times$  & $\surd$  & 6.9 \\
$\surd$ & $\surd$  & $\times$ & 88.3 \\
$\surd$ & $\times$  & $\surd$ & 15.2 \\
$\times$ & $\surd$  & $\surd$ & 87.6 \\
$\surd$ & $\surd$ & $\surd$ &  \textbf{89.1} \\
\hline
\end{tabular}
\end{center}
\caption{Effect of tasks on the accuracy of classification on the ModelNet40.} 
\label{table5}
\end{table}

Finally, we report some of our failed experiments:
\begin{itemize}
\item We tried K-Means++ \cite{Arthur_2007_SIAM} to warm-start the cluster centroids. 
We did not observe any significant improvement  over the randomly selected centroids.

\item We tried soft parameter sharing between the decoder and classifier models. We observed that this destabilizes the model and hence we isolated them.

\item Similar to \cite{Wang_2018_ArXiv}, we tried stacking more graph convolution 
layers and recomputing the input adjacency to each layer based on the feature space 
of its predecessor layer. We observed that this has an adverse effect on both 
classification and segmentation tasks.
\end{itemize}

\section{Conclusion} \label{conclusion}
We proposed an unsupervised multi-task learning approach to learn point and shape 
features on point clouds which uses three unsupervised tasks including clustering, autoencoding, and self-supervised classification to train a multi-scale graph-based 
encoder. We exhaustively evaluated our model on point  cloud classification and
segmentation benchmarks. The results suggest that the learned features outperform 
prior state-of-the-art models in unsupervised representation learning. For example, 
in ModelNet40 shape classification tasks, our model achieved the state-of-the-art 
(among unsupervised models) accuracy of 89.1\% which is also competitive with 
supervised models. In the ShapeNetPart segmentation task, it achieved mIoU of 77.7 
which is only 8\% less than the state-of-the-art supervised model. For future directions, 
we are planning to: (i) introduce more powerful decoders to enhance the quality of 
the learned features, (ii) investigate the effect of other features such as normals and geodesics, and (iii) adapt the model to perform semantic segmentation tasks too.

{\small
\bibliographystyle{ieee_fullname}
\bibliography{egbib}
}

\end{document}